\documentclass[11pt]{article}
\usepackage{lmodern}
    \PassOptionsToPackage{numbers, compress}{natbib}

\usepackage[utf8]{inputenc} 
\usepackage[T1]{fontenc}    

\RequirePackage[colorlinks,citecolor=blue,urlcolor=blue]{hyperref}

\usepackage[authoryear,round]{natbib}

\usepackage{amsmath,amssymb,amsthm}
\usepackage{amscd}

\usepackage{algorithm, algorithmic}
\usepackage{amsfonts,amssymb}
\usepackage{amsmath}
\usepackage{thmtools, thm-restate}
\usepackage{mathrsfs} 
\usepackage[usenames,dvipsnames]{pstricks}

\usepackage[font=small,labelfont=bf]{caption}
\usepackage{subcaption}

\usepackage{url}            
\usepackage{booktabs}       
\usepackage{amsfonts}       
\usepackage{nicefrac}       
\usepackage{microtype}      

\usepackage{multirow}

\usepackage{marvosym}
\usepackage{url}
\usepackage{xcolor}
\usepackage{arydshln}
\usepackage{wrapfig}

\usepackage[capitalize]{cleveref}

\usepackage{graphicx}

\usepackage[shortlabels]{enumitem}

\usepackage{amsmath,amsfonts,bm}

\def\eqref#1{equation~\ref{#1}}

\def\1{\bm{1}}

\DeclareMathAlphabet{\mathsfit}{\encodingdefault}{\sfdefault}{m}{sl}
\SetMathAlphabet{\mathsfit}{bold}{\encodingdefault}{\sfdefault}{bx}{n}

\newcommand{\E}{\mathbb{E}}

\newcommand{\R}{\mathbb{R}}

\DeclareMathOperator*{\argmax}{arg\,max}
\DeclareMathOperator*{\argmin}{arg\,min}

\renewcommand{\paragraph}[1]{\vspace{1em}\noindent{\bfseries #1}.}

\definecolor{dgreen}{rgb}{0.00,0.49,0.00}
\definecolor{dblue}{rgb}{0,0.08,0.75}
\hypersetup{
    colorlinks = true,
    citecolor=dgreen,
    urlcolor=dblue,
    linkcolor=dblue
}

\crefname{assumption}{Assumption}{Assumptions}
\crefname{equation}{}{}
\Crefname{equation}{Eq.}{Eqs.}
\crefname{figure}{Fig.}{Figs.}
\crefname{table}{Tab.}{Tabs.}
\crefname{section}{Sec.}{Sec.}
\crefname{theorem}{Thm.}{Thm.}
\crefname{lemma}{Lemma}{Lemmas}
\crefname{corollary}{Cor.}{Cor.}
\crefname{example}{Example}{Examples}
\crefname{remark}{Remark}{Remarks}
\crefname{algorithm}{Alg.}{Algorightms}

\crefname{appendix}{Appendix}{Appendices}
\crefname{subappendix}{Appendix}{Appendices}
\crefname{subsubappendix}{Appendix}{Appendices}

\newcommand{\imgnet}{\textsc{ImageNet}}
\newcommand{\mimg}{\textit{mini}\imgnet{}}

\newcommand{\EE}{\mathbb{E}}
\newcommand{\Lagr}{\mathcal{L}}

\newcommand{\X}{{\mathcal{X}}}
\newcommand{\Y}{{\mathcal{Y}}}

\newcommand{\D}{D}

\newcommand{\alg}{\textrm{Alg}}

\renewcommand{\citet}{\cite}

\renewcommand{\paragraph}[1]{\noindent{\bfseries #1.}}

\newcommand{\eqal}[1]{\begin{align}#1\end{align}}

\newcommand{\Tau}{\mathcal{T}}
\newcommand{\embd}{\psi}

\newcommand{\schedule}{{S}}

\usepackage[textwidth=2.0cm, textsize=tiny]{todonotes} 

\providecommand{\nor}[1]{\left\|{#1}\right\|}

\newcommand{\ridge}{\phi}

\newcommand{\algname}{SCROLL}

\crefname{assumption}{Assumption}{Assumptions}
\crefname{equation}{Eq.}{Eqs.}
\crefname{figure}{Fig.}{Fig.}
\crefname{table}{Table}{Tables}
\crefname{section}{Sec.}{Sec.}
\crefname{theorem}{Thm.}{Thm.}
\crefname{lemma}{Lemma}{Lemmas}
\crefname{corollary}{Cor.}{Cor.}
\crefname{example}{Example}{Examples}
\crefname{remark}{Remark}{Remarks}
\crefname{algorithm}{Alg.}{Algorightms}

\crefname{appendix}{Appendix}{Appendices}

\makeatletter
\def\@endtheorem{\endtrivlist}
\makeatother

\usepackage{etoolbox}

\AfterEndEnvironment{restatable}{\noindent\ignorespacesafterend}
\AfterEndEnvironment{theorem}{\noindent\ignorespacesafterend}
\AfterEndEnvironment{remark}{\noindent\ignorespacesafterend}
\AfterEndEnvironment{example}{\noindent\ignorespacesafterend}
\AfterEndEnvironment{assumption}{\noindent\ignorespacesafterend}
\AfterEndEnvironment{lemma}{\noindent\ignorespacesafterend}
\AfterEndEnvironment{proof}{\noindent\ignorespacesafterend}
\AfterEndEnvironment{corollary}{\noindent\ignorespacesafterend}
\AfterEndEnvironment{proposition}{\noindent\ignorespacesafterend}
\AfterEndEnvironment{definition}{\noindent\ignorespacesafterend}

\renewcommand{\paragraph}[1]{\vspace{1em}\noindent{\bfseries #1.}}

\graphicspath{{./imgs/}}

\title{\LARGE\bf Schedule-Robust Online Continual Learning\vspace{1em}}

\author{ Ruohan Wang$^{*,1}$ \\ {\footnotesize\em wang\_ruohan@i2r.a-star.edu.sg} \and  Marco Ciccone$^{*,2}$ \\ {\footnotesize\em marco.ciccone@polito.it} \and  Giulia Luise$^{3}$ \\ {\footnotesize\em  g.luise16@ucl.ac.uk} \\ \and  Andrew Yapp$^{5}$ \\ {\footnotesize\em e0360570@u.nus.edu} \and Massimiliano Pontil$^{3,4}$ \\ {\footnotesize\em  massimiliano.pontil@iit.it} \and  Carlo Ciliberto$^{3}$ \\ {\footnotesize\em c.ciliberto@ucl.ac.uk}  }

\date{}

\begin{document}

\maketitle

\begin{abstract}
A continual learning (CL) algorithm learns from a non-stationary data stream.
The non-stationarity is modeled by some schedule that determines how data is presented over time.
Most current methods make strong assumptions on the schedule and have unpredictable performance when such requirements are not met. A key challenge in CL is thus to design methods robust against arbitrary schedules over the same underlying data, since 
in real-world scenarios schedules are often unknown and dynamic.
In this work, we introduce the notion of \textit{schedule-robustness} for CL and a novel approach satisfying this desirable property in the challenging online class-incremental setting.
We also present a new perspective on CL, as the process of learning a schedule-robust predictor, followed by adapting the predictor using only replay data.
Empirically, we demonstrate that our approach outperforms existing methods on CL benchmarks for image classification by a large margin.
\end{abstract}

\section{Introduction}
\label{sec:intro}
\renewcommand{\thefootnote}{\fnsymbol{footnote}}
\footnotetext[1]{Equal Contribution.}
\renewcommand{\thefootnote}{\arabic{footnote}}
\footnotetext[1]{Institute for Infocomm Research, Agency for Science, Technology and Research (A*STAR), Singapore.} \footnotetext[2]{Politecnico di Torino, Torino, Italy (Work done while at University College London).} \footnotetext[3]{Centre for Artificial Intelligence, Department of Computer Science, University College London, United Kingdom.}\footnotetext[4]{Computational Statistics and Machine Learning Group, Istituto Italiano di Tecnologia, Genova, Italy.}\footnotetext[5]{National University of Singapore (Work done while at A*STAR).} A hallmark of natural intelligence is its ability to continually absorb new knowledge while retaining and updating existing one. Achieving this objective in machines is the goal of continual learning~(CL). Ideally, CL algorithms learn online from a never-ending and non-stationary stream of data, without catastrophic forgetting~\citep{mccloskey1989catastrophic, ratcliff1990connectionist, french1999catastrophic}.

The non-stationarity of the data stream is modeled by some \textit{schedule} that defines what data arrives and how its distribution evolves over time. Two family of schedules commonly investigated are {\itshape task-based}~\citep{de2021continualsurvey} and {\itshape task-free}~\citep{aljundi2019task}. The task-based setting assumes that new data arrives one task at a time and data distribution is stationary for each task. Many CL algorithms~\citep[e.g.,][]{buzzega2020dark, kirkpatrick2017overcoming, hou2019learning} thus train \textit{offline}, with multiple passes and shuffles over task data. The task-free setting does not assume the existence of separate tasks but instead expects CL algorithms to learn \textit{online} from streaming data, with evolving sample distribution~\citep{caccia2022new, shanahan2021encoders}. In this work, we tackle the task-free setting with focus on class-incremental learning, where novel classes are observed incrementally and a single predictor is trained to discriminate all of them~\citep{rebuffi2017icarl}.

Existing works are typically designed for specific schedules, since explicitly modeling and evaluating across all possible data schedules is intractable. Consequently, methods have often unpredictable performance when scheduling assumptions fail to hold~\citep{farquhar2018towards, mundt2022clevacompass, Yoon2020Scalable}. This is a considerable issue for practical applications, where the actual schedule is either unknown or may differ from what these methods were designed for. This challenge calls for an ideal notion of \textit{schedule-robustness}: CL methods should behave consistently when trained on different schedules over the same underlying data.

To achieve schedule-robustness, we introduce a new strategy based on a two-stage approach: 1) learning online a schedule-robust predictor, followed by 2) adapting the predictor using only data from experience replay (ER)~\citep{chaudhryER_2019}. We will show that both stages are robust to diverse data schedules, making the whole algorithm schedule-robust. We refer to it as \textbf{SC}hedule-\textbf{R}obust \textbf{O}nline continua\textbf{L} \textbf{L}earning (\algname{}). Specifically, we propose two online predictors that by design are robust against arbitrary data schedules and catastrophic forgetting. To learn appropriate priors for these predictors, we present a meta-learning perspective~\citep{finn2017model, wang2021role} and connect it to the pre-training strategies in CL~\citep{mehta2021empirical}. We show that pre-training offers an alternative and efficient procedure for learning predictor priors instead of directly solving the meta-learning formulation. This makes our method computationally competitive and at the same time offers a clear justification for adopting pre-training in CL. Finally, we present effective routines for adapting the predictors from the first stage. We show that using only ER data for this step is key to preserving schedule-robustness, and discuss how to mitigate overfitting when ER data is limited.

\paragraph{Contributions} 1) We introduce the notion of schedule-robustness for CL and propose a novel online approach satisfying this key property. 2) We present a meta-learning perspective on CL and connect it to pre-training strategies in CL. 3) Empirically, we demonstrate \algname{} outperforms 
a number of baselines by large margins, and highlight key properties of our method in ablation studies.

\section{Preliminaries and Related Works}
\label{sec:bg}
\begin{figure}[t]
    \centering
    \includegraphics[width=\textwidth]{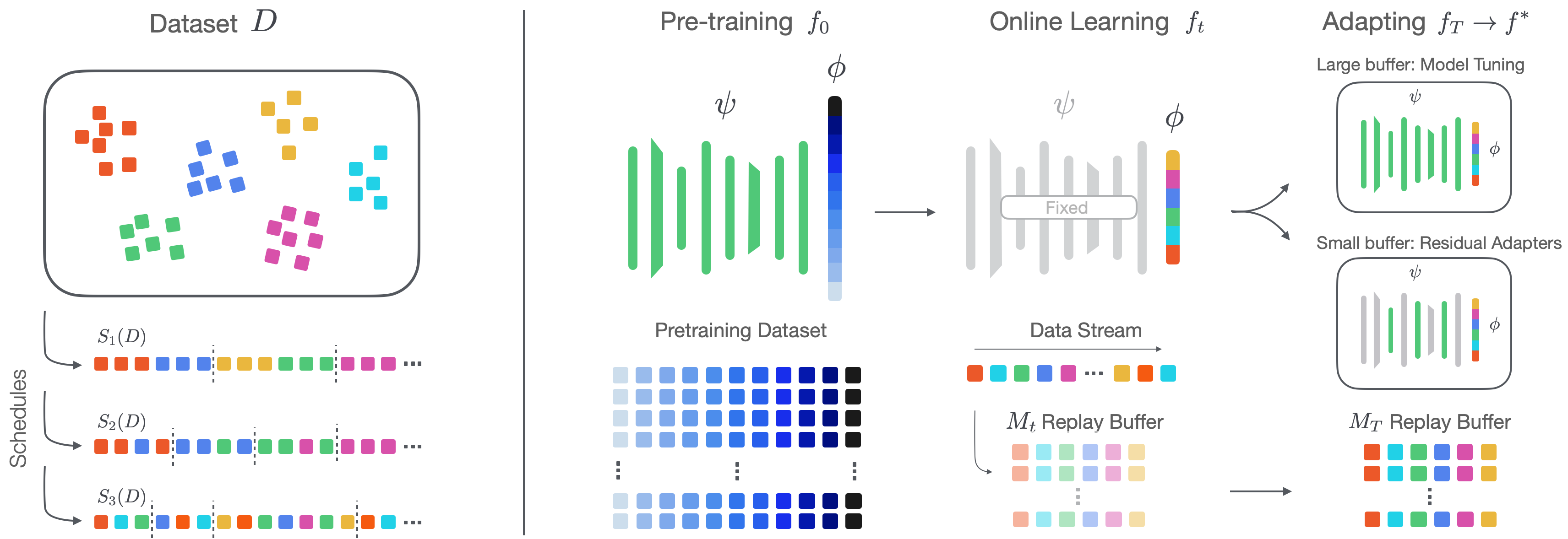}
    \caption{{\bfseries Left.} Illustration of a classification dataset $\D$ streamed according to different schedules (dashed vertical lines identify separate batches).  {\bfseries Right.} Pre-training + the two stages of \algname{}: 1) online learning and store replay samples from data stream, 2) adapting the predictor using the replay buffer (green indicates whether the representation $\embd$ is being updated).}
    \label{fig:tsa_ridge}
\end{figure}

We formalize CL as learning from non-stationary data sequences. A data sequence consists of a dataset $\D=\{(x_i, y_i)\}_{i=1}^N$ regulated by a \textbf{schedule} $\schedule=(\sigma, \beta)$. Applying the schedule $\schedule$ to $\D$ is denoted by $\schedule(\D)\triangleq \beta(\sigma(\D))$, where $\sigma(\D)$ is a specific ordering of $\D$, and $\beta(\sigma(\D)) = \{B_t\}_{t=1}^T$ splits the sequence $\sigma(D)$ into $T$ batches of samples $B_t = \{(x_{\sigma(i)}, y_{\sigma(i)})\}_{i=k_t}^{k_{t+1}}$, with $k_t$ the batch boundaries. Intuitively, $\sigma$ determines the order in which $(x, y) \in \D$ are observed, while $\beta$ determines how many samples are observed at a time. \cref{fig:tsa_ridge}~(Left) illustrates how the same dataset $D$ could be streamed according to different schedules. For example, $\schedule_1(\D)$ in \cref{fig:tsa_ridge} (Left) depicts the standard schedule to split and stream $\D$ in batches of $C$ classes at the time ($C=2$). 

\subsection{Continual Learning}
A CL algorithm learns from $\schedule(\D)$ one batch $B_t$ at a time, iteratively training a predictor $f_t:\mathcal{X}\to\mathcal{Y}$ to fit the observed data. Some formulations assume access to a fixed-size {\itshape replay buffer} $M$, which mitigates forgetting by storing and reusing samples for future training. Given an initial predictor $f_0$ and an initial buffer $M_0$, we define the update rule of a CL algorithm $\alg(\cdot)$ at step $t$ as
\eqal{\label{eq:ft_mt}
    (f_{t},~ M_t) = \alg(B_t, f_{t-1}, M_{t-1}),
}
where the algorithm learns from the current batch $B_t$ and updates both the replay buffer $M_{t-1}$ and predictor $f_{t-1}$ from the previous iteration. 

At test time, the performance of the algorithm is evaluated on a distribution $\pi_{\D}$ that samples $(x,y)$ sharing the same labels with the samples in $\D$. The generalization error is denoted by
\eqal{\label{eq:seq_cont_loss}
    \Lagr(\schedule(\D), f_0, \alg) = \E_{(x, y) \sim \pi_{\D}} \ell(f_T(x), y) \
}
where the final predictor $f_T$ is recursively obtained from \cref{eq:ft_mt} and $f_0$.\footnote{We omitted the initial memory buffer $M_0$ since it is typically empty.} In the following, we review existing approaches for updating $f_t$ and $M_t$, and strategies for initializing $f_0$.

\paragraph{Predictor Update $\boldsymbol{(f_t)}$} Most CL methods learn $f_t$ by solving an optimization problem of the form:
\eqal{\label{eq:update_pred}
    f_{t} = \argmin_{f} \quad \alpha_1\cdot\underbrace{\sum_{(x, y)\in B_t} \ell(f(x), y)}_{\text{current batch loss}} + ~\alpha_2\cdot\underbrace{\sum_{(x, y)\in M_{t-1}} \ell(f(x), y)}_{\text{replay loss}} + ~\alpha_3\cdot \underbrace{R(f, f_{t-1})}_{\text{regularization loss}}
}
where $\alpha_{1,2,3}$ are 
prescribed non-negative weights, $\ell$ a loss function, and $R$ a regularizer. This general formulation for updating $f_t$ recovers replay-based methods such as iCarl~\citep{rebuffi2017icarl} and DER~\citep{buzzega2020dark} 
for specific choices of $\ell$ and $R$.
Moreover, if  $M_{t}{=}\varnothing~\forall t$ (or we set $\alpha_2 {=}0$), the replay loss is omitted and we recover regularization-based methods~\citep[e.g][]{kirkpatrick2017overcoming, li2017learning, yu2020semantic} that update selective parameters of $f_{t-1}$ to prevent forgetting.

\paragraph{Replay Buffer Update  $\boldsymbol{(M_t)}$} In replay-based methods, $\alg(\cdot)$ must also define a \textit{buffering strategy} that decides what samples to store in $M_t$ for future reuse, and which ones to discard from a full buffer. Common strategies include exemplar selection~\citep{rebuffi2017icarl} and random reservoir sampling~\citep{vitter1985random}, with more sophisticated methods like gradient-based sample selection~\citep{aljundi2019gradient} and meta-ER~\citep{riemer2018learning}. As in \cref{eq:update_pred}, replay-based methods typically mix the current data with replay data for predictor update~\citep[e.g][]{aljundi2019gradient, riemer2018learning, caccia2022new}. In contrast, \cite{prabhu2020gdumb} learns the predictor using only replay data. We will adopt a similar strategy and discuss how it is crucial for achieving schedule-robustness.

\paragraph{Predictor Initialization $\boldsymbol{(f_0)}$} The initial predictor $f_0$ in \cref{eq:seq_cont_loss} represents the prior knowledge available to CL algorithms, before learning from sequence $S(D)$. Most methods are designed for randomly initialized $f_0$ with no prior knowledge~\citep[e.g.,][]{rebuffi2017icarl, gupta2020look, prabhu2020gdumb, kirkpatrick2017overcoming}. However, this assumption may be overly restrictive for several applications (e.g., vision-based tasks like image classification), where available domain knowledge and data can endow CL algorithms with more informative priors than random initialization. We review two strategies for predictor initialization relevant to this work.

\textit{Initialization by Pre-training.} One way for initializing $f_0$ is to pre-train a representation on data related to the CL task (e.g., ImageNet for vision-based tasks) via either self-supervised learning~\citep{shanahan2021encoders} or multi-class classification~\citep{mehta2021empirical, wang2022learning, wu2022class}. \cite{boschini2022transfer} observed that while pre-training mitigates forgetting, model updates quickly drift the current $f_t$ away from $f_0$, diminishing the benefits of prior knowledge as CL algorithms continuously learn from more data. To mitigate this, \cite{shanahan2021encoders, wang2022learning} keep the pre-trained representation fixed while introducing additional parameters for learning the sequence. In this work, we will offer effective routines for updating the pre-trained representation and significantly improve test performance.

\textit{Initialization by Meta-Learning.} Another approach for initializing $f_0$ is meta-learning~\citep{hospedales2021meta}. Given the CL generalization error in \cref{eq:seq_cont_loss}, we may learn $f_0$ by solving the meta-CL problem below,
\eqal{\label{eq:meta-learning-risk}
f_0 = \argmin_f ~~ \EE_{(\D, \schedule) \sim \Tau} ~\Lagr(\schedule(\D), f, \alg)
}
where $\Tau$ is a meta-distribution over datasets $\D$ and schedules $\schedule$. For instance, \cite{javed2019meta} set $\alg(\cdot)$ to be MAML~\citep{finn2017model} and observed that the learned $f_0$ encodes sparse representation to mitigate forgetting. However, directly optimizing \cref{eq:meta-learning-risk} is computationally expensive since the cost of gradient computation scales with the size of $\D$. To overcome this, we will leverage \cite{wang2021role} to show that meta-learning $f_0$ is analogous to pre-training for certain predictors, which provides a much more efficient procedure to learn $f_0$ without directly solving \cref{eq:meta-learning-risk}.

\subsection{Schedule-Robustness}
The performance of many existing CL methods implicitly depends on the data schedule, leading to unpredictable behaviors when such requirements are not met~\citep{farquhar2018towards, Yoon2020Scalable,mundt2022clevacompass}. To tackle this challenge, we introduce the notion of schedule-robustness for CL. Given a dataset $\D$, we say that a CL algorithm is \textbf{schedule-robust} if
\eqal{\label{eq:seq_agn}
    \Lagr(\schedule_1(\D), f_0, \alg) \approx \Lagr(\schedule_2(\D), f_0, \alg), \qquad \forall \schedule_1, \schedule_2 ~~\text{schedules}.
}
\Cref{eq:seq_agn} captures the idea that CL algorithms should perform consistently across arbitrary schedules over the same dataset $\D$. We argue that achieving robustness to different data schedules is a key challenge in real-world scenarios, where data schedules are often unknown and possibly dynamic. CL algorithms should thus carefully satisfy \cref{eq:seq_agn} for safe deployment.

We note that schedule-robustness is a more general and stronger notion than order-robustness from \cite{Yoon2020Scalable}. Our definition applies to online task-free CL while order-robustness only considers offline task-based setting. We also allow arbitrary ordering for individual samples instead of task-level ordering. In the following, we will present our method and show how it achieves schedule-robustness.

\section{Method}
We present \textbf{SC}hedule-\textbf{R}obust \textbf{O}nline continua\textbf{L} \textbf{L}earning (\algname{}) as a two-stage process: 1) learning online a schedule-robust predictor for CL, followed by 2) adapting the predictor using only replay data. In the first stage, we consider two schedule-robust predictors and discuss how to initialize them, motivated by the meta-CL perspective introduced in \cref{eq:meta-learning-risk}. In the second stage, we tackle how to adapt the predictors from the first stage with ER and the buffering strategy. We will show that \algname{} satisfies schedule-robustness {\itshape by construction}, given that optimizing a CL algorithm against all possible schedules as formulated in \cref{eq:seq_agn} is clearly intractable.

\subsection{Schedule-robust Online Predictor}
\label{sec:method_ci_meta}
We model the predictors introduced in \cref{eq:ft_mt} as the composition $f=\ridge\circ\embd$ of a feature extractor $\embd:\X\to\R^m$ and a classifier $\ridge:\R^m\to\Y$. In line with recent meta-learning strategies~\citep[e.g.][]{bertinetto2018meta, raghu2019rapid}, we keep $\embd$ fixed during our method's first stage while only adapting the classifier $\ridge$ to learn from data streams. We will discuss how to learn $\embd$ in \cref{sec:method_pre-train}. 

A key observation is that some choices of $\ridge$, such as Nearest Centroid Classifier (NCC)~\citep{salakhutdinov2007learning} and Ridge Regression~\citep{kailath2000linear}, are schedule-robust by design.

\paragraph{Nearest Centroid Classifier (NCC)} NCC classifies a sample $x$ by comparing it to the learned \textit{``prototypes''} $c_y$ for each class $X^y \triangleq\{x | (x, y) \in D\}$ in the dataset,
\eqal{\label{eq:proto}
    f(x) = \argmin_{y} \nor{\embd(x) - c_y}^2_2 \quad \textrm{ where } \quad c_y = \frac{1}{n}\sum_{x\in X^y} \embd(x).
}
The prototypes $c_y$ can be learned online: given a new $(x, y)$, we update only the corresponding $c_y$ as
\eqal{\label{eq:proto_seq}
    c_y^{\text{new}} = \frac{n_y \cdot c_y^{\text{old}} + \embd(x)}{n_y+1},
}
where $n_y$ is the number of observed samples for class $y$ so far. We note that the prototype for each class is invariant to any ordering of $D$ once the dataset has been fully observed. Since $\embd$ is fixed, the resulting predictor in \cref{eq:proto} is schedule-robust and online. Further, $f$ is unaffected by catastrophic forgetting by construction, as keeping $\embd$ fixed prevents forgetting while learning each prototype independently mitigates cross-class interference~\citep{hou2019learning}.

\paragraph{Ridge Regression} Ridge Regression enjoys the same desirable properties of NCC, being schedule-robust and unaffected by forgetting (see \cref{sec:app_ls} for a full derivation). Let $\textrm{OneHot}(y)$ denote the one-hot encoding of class $y$, we obtain the predictor $f$ as the one-vs-all classifier
\eqal{
\begin{split}
    \label{eq:ridge_reg}
    f(x) & = \argmax_{y}~ w_y^\top\embd(x), \qquad \textrm{where}\\
     & W^* = [w_1 \dots w_K] = \argmin_{W}\frac{1}{|D|}\sum_{(x,y)\in D}~~\nor{W^\top\embd(x) - \textrm{OneHot}(y)}^2 + \lambda\nor{W}^2.
\end{split}
}
Ridge Regression admits online updates via recursive least squares~\citep{kailath2000linear}: given $(x, y)$,
\eqal{
\begin{split}
\label{eq:ridge_seq}
      w_z  = (& A^{\text{new}}  + \lambda I)^{-1}c_z^{\text{new}} \qquad \forall z\in\Y, \qquad \textrm{where} \\ 
      & A^{\text{new}} = A^{\text{old}} + \embd(x)\embd(x)^\top \qquad \textrm{ and } \qquad c_z^{new} = \begin{cases}
        c_z^{\text{old}} + \embd(x) & \text{if } z=y\\
        c_z^{\text{old}} & \text{otherwise}.
    \end{cases}
\end{split}
}
Here, $A$ denotes the covariance matrix of all the samples observed so far during a schedule, while $c_z$ is the sum of the embeddings $\embd(x)$ of samples $x$ in class $z$.

\subsection{Predictor Initialization}
\label{sec:method_pre-train}
Within our formulation, initializing $f_0$ consists of learning a suitable feature representation $\embd$. As discussed in \cref{sec:bg}, we could learn $f_0 = \phi_0 \circ\embd$ by solving the meta-CL problem in \cref{eq:meta-learning-risk} (with $\phi_0$ a classifier for the meta-training data to be discarded after initialization, see \cref{fig:tsa_ridge} (Right)). As shown in \cref{sec:method_ci_meta}, if we set $\alg(\cdot)$ in \cref{eq:meta-learning-risk} to be the online versions of NCC or Ridge Regression, the resulting predictor $f_T$ becomes invariant to any schedule over $\D$. This implies that {\itshape the meta-CL problem is equivalent to a standard meta-learning one where all data in $\D$ are observed at once}, namely
\eqal{\label{eq:meta-risk-reduced}
    f_0 = \argmin_f ~~ \EE_{\D\sim\Tau} ~\EE_{(x,y)\sim\pi_D} \ell(f_\D(x),y) \qquad \textrm{where} \qquad f_D = \alg(\D,f).
}
For example, setting $\alg(\cdot)$ to NCC recovers exactly the well-known ProtoNet~\citep{snell2017prototypical}.

\paragraph{Meta-learning and pre-training} 
Meta-learning methods for solving \cref{eq:meta-risk-reduced} are mostly designed for low data regimes with small $D$ while in contrast, CL has to tackle long sequences (i.e. $\D$ is large). In these settings, directly optimizing \cref{eq:meta-risk-reduced} quickly becomes prohibitive since the cost of computing gradients scales linearly with the size of $\D$. Alternatively, recent works in meta-learning strongly support learning $f_0$ via standard multi-class classification~\citep{tian2020rethinking, wang2021role}. In particular, \cite{wang2021role} showed that the cross-entropy loss in multi-class classification is an upper bound to meta-learning $f_0$ in \cref{eq:meta-risk-reduced}. Classification thus offers a theoretically sound estimator for meta-learning. As a direct consequence, our meta-learning perspective on initializing $f_0$ provides a principled justification for exploiting pre-training in CL. Pre-training is computationally efficient, since gradients can be computed on small batches rather than the entire $\D$ in \cref{eq:meta-risk-reduced}. Extensive evidence also suggests that pre-training often yields more robust $f_0$ than meta-learning~\citep{el2022lessons}.

\paragraph{Learning $\boldsymbol{f_0}$ in \algname{}} Given the  reasoning above, we propose to not tackle \cref{eq:meta-risk-reduced} directly but rather learn $f_0$ by multi-class classification. Following the standard practices from meta-learning, we ensure the meta-training data to be disjoint from the sequences we run the CL algorithm on, such that $\alg(\cdot)$ is indeed learning novel data during CL. For instance, in \cref{sec:exp-cifar} we will use $f_0$ trained on Meta-Dataset~\citep{triantafillou2019meta} to perform CL on CIFAR datasets, which do not overlap. 



\subsection{Predictor Adaptation with Experience Replay}
\label{sec:method_replay}
The predictor $f_T$ obtained via online learning is schedule-robust and readily available for classification. However, it shares the same feature extractor $\embd$ with $f_0$ and may therefore suffer from the distribution shift between the pre-training data and $\D$. To address this problem we propose to further adapt $f_T$ into a $f^*$ by using Experience Replay (ER). Our key insight is to consider buffering strategies that yield similar replay buffers, regardless of the schedule over $\D$, hence keeping $f^*$  schedule-robust.

\paragraph{Buffering strategy} We adopt the \textit{exemplar buffering} from \cite{rebuffi2017icarl}. The mehod uses greedy moment matching to select the same number of samples for each class independently, according to how well the mean embedding (via $\embd$) of the selected samples approximates the mean embedding of all points in the class. The greedy selection provides a total ordering for all points in each class, which is used to determine which samples to store or discard. This step filters out outliers and ensures that $M_T$ contains representative samples from each class.

This process aims to keep each $M_t$ balanced class-wise, hence the final $M_T$ attains a similar class distribution after $\D$ is fully observed, irrespective of the specific schedule (we refer to \cref{sec:app_schd} for empirical evidence in support of this observation). In particular, when we present one entire class at a time (namely no $X^y$ is split across multiple batches) such as in few shot CL~\citep{antoniou2020defining}, {\itshape exemplar buffering is invariant to the order in which classes are presented}. Thus, adapting $f_T$ with $M_T$ yields a schedule-robust $f^*$. Below we consider two adaptation strategies.



\paragraph{Small buffer size: residual adapters} When $M$ has limited capacity (e.g., 200 or 500 in our experiments), there may be insufficient data for adapting all parameters in $f_T$. To avoid overfitting, we introduce residual adapters inspired by task-specific adapters~\citep{li2022cross}. Concretely, we insert residual layers (e.g. residual convolutions) to each level of $\embd$. Then, while keeping $\embd$ fixed, we jointly optimize the new layers and the classifier $\ridge$ using a cross-entropy loss (see \cref{fig:tsa_ridge} (Right)). 


\paragraph{Large buffer size: model tuning} For larger buffers, we empirically observed that updating all parameters of $f_T$ is more effective. This is not surprising given the added representational power from the full model. We will also demonstrate that the threshold for switching from residual adapters to full-model tuning appears consistent and may be determined a priori.

\subsection{Practical Algorithm}
\label{sec:method_alg}

\begin{wrapfigure}{R}{0.47\textwidth}
    \vspace*{-25pt}
    \begin{minipage}{0.45\textwidth}
        \begin{algorithm}[H]
           \caption{\algname{}\label{alg:meta-CI}}
        \begin{algorithmic}
            \small
            \STATE {\bfseries Input:} Pre-trained $\embd$, CL sequence $\schedule(D)$, Buffer $M_0 = \varnothing$, buffer max size $m$.
            \vspace{0.5em}
            \STATE $f_T, M_T = ${ \sc Online-CL}$(\embd, S(D), M_0)$
            \STATE {\bfseries If} $|M_T| \le m$:
            \STATE \quad $f^* =  \textrm{ResidualAdapt}(f_T, M_T)$
            \STATE {\bfseries Else:}
            \STATE \quad $f^* = \textrm{FullAdapt}(f_T, M_T)$
            \STATE {\bfseries return} $f^*$
            \vspace{0.5em}
            \STATE{\bfseries def}~ {\sc Online-CL} $(\embd, \schedule(\D), M_0)$
            \STATE\quad {\bfseries For} $B_t$ in $S(\D)$:
            \STATE\qquad Update $M_i$ (exemplar strategy, \cref{sec:method_replay})
            \STATE\qquad Update statistics $c_y, A$ using \cref{eq:proto_seq} or
            \cref{eq:ridge_seq}.
            \STATE\quad Compute classifier $\ridge$ via \cref{eq:proto} or \cref{eq:ridge_seq}.
            \STATE\quad {\bfseries return} $f_T=\ridge\circ\embd$ and $M_T$
        \end{algorithmic}
        \end{algorithm}
    \vspace*{-35pt}
    \end{minipage}
\end{wrapfigure}
We present the complete algorithm in \cref{alg:meta-CI} (see also \cref{fig:tsa_ridge}): We first initialize $f_0$ with a pre-trained $\embd$ via multi-class classification (\cref{sec:method_pre-train}). As \algname{} observes a sequence $\schedule(D)$ online, it updates $f_t$ by storing the necessary statistics for computing the classifier $\ridge$ (\cref{sec:method_ci_meta}), and the replay buffer $M_t$ (\cref{sec:method_replay}). At any time $T$ (e.g. the end of the sequence), a model $f^*$ can be obtained by adapting $f_T$ using $M_T$.


We highlight several properties of \algname{}: 1) it is schedule-robust; 2) it is designed for class-incremental learning with a single classifier for all observed classes; 3) it meets the definition of online CL~\citep{chaudhry2018efficient, mai2022online}, where the data stream is observed once, and only the replay data is reused.
\newpage
\paragraph{Intermediate predictors} Existing CL methods can produce intermediate predictors for partial sequences, such as after observing each class. \algname{} can similarly do so \textit{on demand}: whenever a predictor is required, \algname{} could compute the current predictor $f_t=\ridge\cdot\embd$ with the stored statistics, followed by adapting $f_t$ on the current buffer $M_t$. The key difference from existing methods is that we will \textit{always begin the adaptation from the initial $\embd$ to preserve schedule-robustness}. This process also explains how forgetting occurs in our method: increased classification difficulty from learning new classes and fewer samples per class for ER due to buffer capacity constraint.

\paragraph{Memory-free \algname{}} We highlight that the $f_T$ returned by the Online-CL routine in \cref{alg:meta-CI}, is already a valid and deterministic predictor for $\D$. Hence, if we have constraints  (e.g., privacy concerns~\cite{shokri2015privacy}) or memory limitations, \algname{} can be memory-free by not using ER. We empirically investigate such strategy in our ablation studies \cref{sec:exp_ab}.

\section{Experiments}
\label{sec:method}
We consider class-incremental learning and compare \algname{} primarily to online CL methods, including GDumb~\citep{prabhu2020gdumb}, MIR~\citep{aljundi2019mir}, ER-ACE and ER-AML~\citep{caccia2022new}, and SSIL~\citep{ahn2021ss}. As shown by~\citep{buzzega2020dark}, online methods often perform poorly due to the limited number of model updates. To disentangle forgetting from underfitting in the online setting, we further compare to recent offline methods for completeness, including ER~\citep{riemer2018learning}, BIC~\citep{wu2019large}, DER and DER++~\citep{buzzega2020dark}. We recall that the conventional offline setting allows multiple passes and shuffles over data for each task~\citep{de2021continualsurvey}. We differentiate our method by the classifier used, namely \algname{}~(NCC) and \algname{}~(Ridge). The source code for our method will be made available shortly.

\paragraph{Setup} We first consider sequential CIFAR-10 and sequential CIFAR-100 benchmarks. We use a ResNet18 initial predictor trained on Meta-Dataset using multi-class classification~\citep{li2021universal}. Following the convention of meta-learning, Meta-Dataset has no class overlap with CIFAR datasets. This ensures that CL algorithms must learn about the new classes during CL. We also evaluate \algname{} on \mimg{} as another sequential learning task to highlight that even a weakly initialized $f_0$ is very beneficial. Following \cite{javed2019meta}, all methods are allowed to learn an initial predictor using the meta-training set of \mimg{} (38400 samples over 64 classes), and perform CL on the meta-testing set (20 novel classes).

\paragraph{Baseline methods} For fairness, {\itshape all baseline methods use the same pre-trained feature extractor}. We also performed extensive hyper-parameter tuning for all baseline methods such that they perform well with pre-training (see \cref{sec:app_base_tune}). Lastly, we standardize that buffer size reported in all experiments as the total buffer for all classes, not buffer per class.

\subsection{Class-incremental Learning on CIFAR Datasets}\label{sec:exp-cifar}
For CIFAR datasets, we first follow a standard schedule from \cite{buzzega2020dark, wu2019large}: CIFAR-10 is divided into 5 splits of 2 classes each while CIFAR-100 into 10 splits of 10 classes each. At each split, this schedule abruptly changes from one data distribution to another.

\begin{table}[tb]
    \centering
    \scriptsize
    \begin{tabular}{ll|ccc|ccc}
        & & \multicolumn{3}{c|}{\textbf{CIFAR-10 5-split}} & \multicolumn{3}{c}{\textbf{CIFAR-100 10-split}}\\
        \midrule
        \multicolumn{2}{c|}{Joint Training (i.i.d.)} & \multicolumn{3}{c|}{$94.2$ \tiny{$\pm 1.6$}} & \multicolumn{3}{c}{$79.3$ \tiny{$\pm 0.4$}}\\
        \midrule
        \multicolumn{2}{c|}{\textbf{CL Algorithm}} & $|M|=200$ & $|M|=500$ & $|M|=2000$ & $|M|=200$ & $|M|=500$ & $|M|=2000$\\
        \midrule
        \multirow{4}{*}{\rotatebox[origin=c]{90}{\textsc{Offline}}} 
        & ER   & $67.9$ \tiny{$\pm1.1$} & $77.3$ \tiny{$\pm 0.4$} & $86.8$ \tiny{$\pm 0.4$} & $22.9$ \tiny{$\pm 0.6$} & $33.2$ \tiny{$\pm 0.6$} & $53.1$ \tiny{$\pm 0.3$} \\
        & BIC & $\mathbf{77.4}$ \tiny{$\pm 2.2$} & $\mathbf{88.6}$ \tiny{$\pm 5.3$} & $\mathbf{90.4}$ \tiny{$\pm 0.5$} & $\mathbf{34.9}$ \tiny{$\pm 1.6$} & $\mathbf{45.2}$ \tiny{$\pm 1.9$} & $57.5$ \tiny{$\pm0.7$}  \\
        & DER   & $69.7$ \tiny{$\pm1.7$} & $78.2$ \tiny{$\pm 0.8$} & $86.6$ \tiny{$\pm 0.7$} & $26.2$ \tiny{$\pm 0.8$} & $38.1$ \tiny{$\pm 2.1$} & $55.3$ \tiny{$\pm 0.6$} \\
        & DER++ & $76.8$ \tiny{$\pm1.7$} & $81.6$ \tiny{$\pm 2.2$} & $86.8$ \tiny{$\pm 0.8$} & $28.5$ \tiny{$\pm 0.7$} & $42.7$ \tiny{$\pm 1.1$} & $\mathbf{59.1}$ \tiny{$\pm 0.5$} \\
        \midrule
        \multirow{5}{*}{\rotatebox[origin=c]{90}{\textsc{Online}}}
        & GDumb  & $73.6$ \tiny{$\pm 1.9$} & $81.2$ \tiny{$\pm 0.9$} & $87.3$ \tiny{$\pm 0.3$} & $21.9$ \tiny{$\pm 1.3$} & $39.0$ \tiny{$\pm 1.0$} & $59.8$ \tiny{$\pm 0.2$} \\
        & ER-ACE & $77.8$ \tiny{$\pm 0.9$} & $81.4$ \tiny{$\pm 0.6$} & $84.7$ \tiny{$\pm 0.8$} & $40.3$ \tiny{$\pm 1.2$} & $49.0$ \tiny{$\pm 0.9$} & $56.5$ \tiny{$\pm 1.0$} \\
        & ER-AML & $69.9$ \tiny{$\pm 2.9$} & $75.3$ \tiny{$\pm 0.7$} & $80.6$ \tiny{$\pm 0.6$} & $32.4$ \tiny{$\pm 0.5$} & $41.9$ \tiny{$\pm 1.0$} & $51.0$ \tiny{$\pm 1.2$} \\
        & SSIL   & $68.9$ \tiny{$\pm 1.4$} & $70.8$ \tiny{$\pm 1.4$} & $73.2$ \tiny{$\pm 1.7$} & $46.6$ \tiny{$\pm 1.0$} & $53.9$ \tiny{$\pm 0.6$} & $59.2$ \tiny{$\pm 0.5$} \\
        & MIR & $55.2$ \tiny{$\pm 1.5$} & $68.7$ \tiny{$\pm 1.3$}  & $82.0$ \tiny{$\pm 1.0$} & $23.2$ \tiny{$\pm 0.8$} & $31.9$ \tiny{$\pm 0.5$} & $47.5$ \tiny{$\pm 1.1$} \\
        \hdashline[1pt/1pt]\noalign{\vskip 1.0ex}
        \multirow{2}{*}{\rotatebox[origin=c]{90}{(ours)}}
        & \algname{} (NCC)   & $81.5$ \tiny{$\pm 0.1$} & $84.6$ \tiny{$\pm 0.1$} & $88.4$ \tiny{$\pm 0.1$} & $48.3$ \tiny{$\pm 0.1$} & $56.4$ \tiny{$\pm 0.2$} & $66.6$ \tiny{$\pm 0.2$} \\
        & \algname{} (Ridge) & $\mathbf{84.0}$ \tiny{$\pm 0.2$} & $\mathbf{86.3}$ \tiny{$\pm 0.2$} & $\mathbf{89.4}$ \tiny{$\pm 0.2$} & $\mathbf{59.9}$ \tiny{$\pm 0.1$} & $\mathbf{61.5}$ \tiny{$\pm 0.1$} & $\mathbf{68.0}$ \tiny{$\pm 0.2$}\\
    \end{tabular}
    \caption{Class-incremental classification accuracy on sequential CIFAR-10\,/\,100. Joint training accuracy obtained by training on all classes with standard supervised learning. Best online/offline methods in bold.}
    \label{tab:main_comp}
\end{table}



\cref{tab:main_comp} shows that \algname{}~(Ridge) outperforms all baselines by large margins on CIFAR-100 for all buffer sizes. For CIFAR-10, \algname{} only lags behind BIC for buffer sizes 500 and 2000 while outperforming the rest. However, we emphasize that BIC is an offline method and not directly comparable to ours. In addition, we will demonstrate next that most baseline methods including BIC have unreliable performance when the schedule changes.

\begin{figure}[t]
  \centering
  \includegraphics[width=\textwidth]{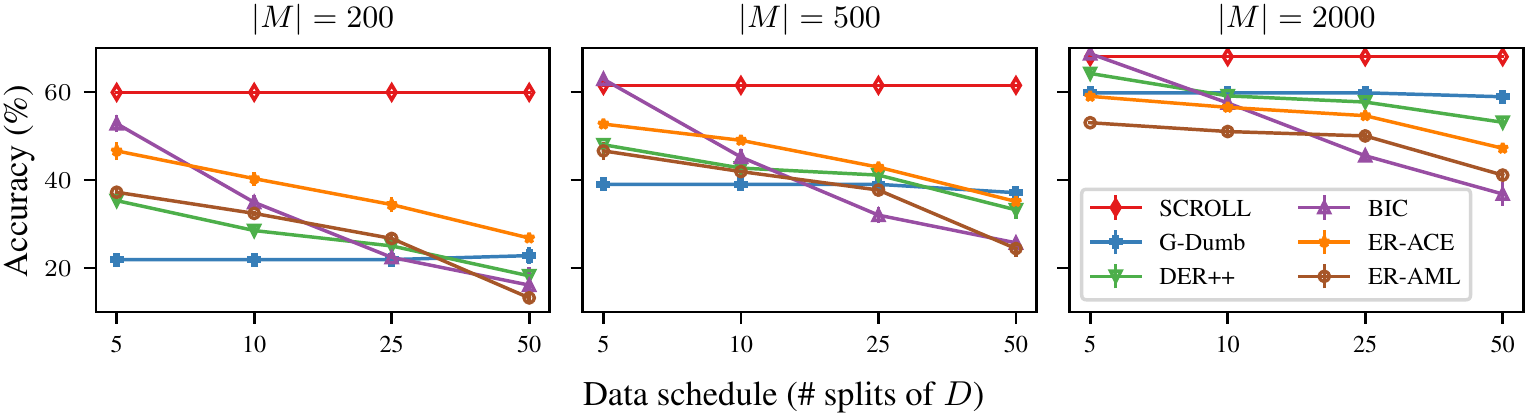}
  \caption{Robustness of \algname{} vs offline/online baselines to different schedules of CIFAR-100.}
  \label{fig:splits_compare}
\end{figure}

\paragraph{Varying Schedules} We now consider the effect of using schedules for CIFAR-100 with splits presenting 20\,/\,10\,/\,4\,/\,2 classes at a time (respectively 5\,/\,10\,/\,25\,/\,50-splits). \cref{fig:splits_compare}, evaluates the best performing baselines from \cref{tab:main_comp} under these schedules. We observe that \algname{} performs consistently across all settings and outperforms all baselines by a large margin, with BIC performing competitively only under the 5-split. Additionally, we clearly see that the accuracy of both online and offline methods drops drastically as the dataset split increases. This further validates \cite{Yoon2020Scalable, mundt2022clevacompass} who observed that existing methods are brittle to schedule variations. The only exception is GDumb, which only uses replay data and is thus (mostly) robust against schedule changes, according to our discussion in \cref{sec:method_replay}. We note, however, that GDumb performs poorly in some ``worst-case'' regimes (see \cref{sec:app_buffer}). 
Additional results are reported in \cref{sec:app_add_split,sec:app_comp_ens}. including a comparison with  ~\citep{shanahan2021encoders} who introduces an ensemble CL method to tackle schedules variations, and a challenging Gaussian schedule for empirical evaluation. 


\subsection{Class-incremental Learning on \mimg{}}
\label{sec:exp_mimg}

%
\begin{wraptable}{R}{0.47\textwidth}
    \centering
    \scriptsize
    \begin{tabular}{llcc}
        & & \multicolumn{2}{c}{\textbf{\mimg{} 10-split}}\\
        \multicolumn{2}{c}{Joint Training (i.i.d.)} & \multicolumn{2}{c}{$89.2$ \tiny{$\pm 0.4$}}\\
        \midrule
        \multicolumn{2}{c}{\textbf{CL Algorithm}} & $|M|=200$ & $|M|=500$ \\
        \midrule
        \multirow{4}{*}{\rotatebox[origin=c]{90}{\textsc{Offline}}}
        & BIC   & $64.6$ \tiny{$\pm 3.3$} & $73.4$ \tiny{$\pm 1.6$}\\
        & ER    & $63.7$ \tiny{$\pm 1.0$} & $73.6$ \tiny{$\pm 0.6$}\\
        & DER   & $67.3$ \tiny{$\pm 1.1$} & $75.2$ \tiny{$\pm 0.8$}\\
        & DER++ & $68.8$ \tiny{$\pm 1.0$} & $78.0$ \tiny{$\pm 0.6$}\\
        \midrule
        \multirow{5}{*}{\rotatebox[origin=c]{90}{\textsc{Online}}}
        & GDumb  & $65.6$ \tiny{$\pm 1.3$} & $74.5$ \tiny{$\pm 0.5$}\\
        & ER-ACE & $57.5$ \tiny{$\pm 1.4$} & $64.4$ \tiny{$\pm 0.6$}\\
        & ER-AML & $52.4$ \tiny{$\pm 2.2$} & $58.4$ \tiny{$\pm 1.7$}\\
        & SSIL   & $57.0$ \tiny{$\pm 1.5$} & $63.3$ \tiny{$\pm 1.6$}\\
        & MIR    & $48.4$ \tiny{$\pm 2.8$} & $67.3$ \tiny{$\pm 2.9$}\\
        \hdashline[1pt/1pt]\noalign{\vskip 1.0ex}
        \multirow{2}{*}{\rotatebox[origin=c]{90}{(ours)}}
        & \algname{}~(NCC) & $72.8$ \tiny{$\pm 0.2$} & $76.0$ \tiny{$\pm 0.1$}\\
        & \algname{}~(Ridge) & $\mathbf{80.9}$ \tiny{$\pm 0.1$} & $\mathbf{82.0}$ \tiny{$\pm 0.1$}\\ 
    \end{tabular}
    \caption{Classification accuracy on sequential \mimg{} 10-split.}
    \label{tab:main_comp_mini}
\end{wraptable}

We evaluate \algname{} on \mimg{} 10-split following the setup described at the start of this section. We stress that all methods perform pre-training on the small meta-training set and then carry out CL on the meta-test set. \cref{tab:main_comp_mini} reports the results of our experiments. We note that \algname{}~(Ridge) outperforms all baseline methods with noticeable improvements. This suggests that our method does not require pre-training on massive datasets and could work well with limited data. It is also clear that the Ridge Regression classifier plays an important role in the final performance, while \algname{} (NCC) only achieves accuracy comparable to the baselines. We further investigate this phenomenon in the following ablation studies.
\newpage
\subsection{Ablation Studies}
\label{sec:exp_ab}
We now provide additional insights on \algname{} by studying the impact of the two main components of the algorithm, namely the use of linear online learning estimators to initialize the model adaptation and the size of the replay buffer. 

\paragraph{Effects of classifier initialization}
In our formulation of ER, the role of NCC or Ridge Regression is to initialize the classifier $\ridge$ at $f_T$ before adapting $f_T$ towards $f^*$. In principle, one could skip this online learning step (i.e. Online-CL routine in \cref{alg:meta-CI}), initialize $\ridge$ in $f_T$ randomly, and adapt $f_T$. \cref{tab:comp_init} compares these options for initializing the classifier\footnote{We will discuss in \cref{sec:app_model} how to implement \cref{eq:proto} as a standard linear layer}.

The results suggest that how classifier $\ridge$ is initialized at $f_T$ drastically affect the performance of final predictor $f^*$. In particular, the performance of $f_T$ correlates strongly with the performance of $f^*$: Ridge Regression has the best predictor both before and after adaptation. In contrast, randomly initialized $\ridge$ does not learn from data streams, and the resulting $f^*$ falls short of the other two initializers. Crucially, the results suggest that online learning described in \cref{sec:method_ci_meta} is vital for fully exploiting the pre-training.
We also remark that while randomly initialized $\ridge$ resembles GDumb, there are still crucial differences between them, including buffering strategy, model architecture and training routine. In particular, we study in depth the role of the buffering strategy in \cref{sec:app_buffer}.

\begin{wraptable}{R}{0.46\textwidth}
\setlength{\tabcolsep}{2pt}
    \centering
    \scriptsize
    \begin{tabular}{c|cc|cc}
         & \multicolumn{2}{c|}{\textbf{CIFAR-100 10-split}} & 
         \multicolumn{2}{c}{\textbf{\mimg{} 10-split}} \\
         \midrule
         $\boldsymbol{\ridge}$ \textbf{Init} & \multicolumn{2}{c|}{$f_T ~~ \rightarrow  ~~ f^*$} & \multicolumn{2}{c}{$f_T ~~  \rightarrow ~~ f^*$}\\
         \midrule
         Random & $1.2$ \tiny{$\pm 0.3$} & $49.0$ \tiny{$\pm 0.2$} & $4.9$ \tiny{$\pm 0.3$} & $75.3$ \tiny{$\pm 0.8$}\\
         NCC & $40.4$ \tiny{$\pm 0$} & $56.4$ \tiny{$\pm 0.2$} & $65.3$ \tiny{$\pm 0$} & $76.0$ \tiny{$\pm 0.1$}\\
         Ridge & $57.1$ \tiny{$\pm 0$} & $61.5$ \tiny{$\pm 0.1$} & $80.6$ \tiny{$\pm 0.2$} & $82.0$ \tiny{$\pm 0.1$}
    \end{tabular}
    \caption{Classification accuracy with different classifier initialization for $|M|=500$.}
    \label{tab:comp_init}
\end{wraptable}

\paragraph{Effects of replay buffer size} We study the impact of changing replay buffer capacity for \algname{}. We focus on \algname{}~(Ridge), given its better performance over the NCC variant. \cref{tab:buffer_size} reports how buffer capacity affect model performance using either residual adapters or full-model tuning.

We first observe that ER contributes significantly to final performance. Using Ridge Regression, the memory-free $f_T$ only obtains 57\% for CIFAR-100 using Ridge Regression. Even with a limited buffer of 200 (i.e. 2 samples per class), ER adds about 3\% to the final accuracy. When $M=2000$, ER adds more than 10\%. The results suggest that updating the pre-trained representation and the associated classifier is important. When combined with the effects from classifier initialization, the results suggest that all components in our method are important for learning robust predictors. 

\cref{tab:buffer_size} also suggests that full-model tuning causes overfitting when replay data is limited. At $|M|=200$, the final predictor $f^*$ consistently underperforms $f_T$. As the buffer size increases, full-model tuning becomes more effective compared to residual adapters. For CIFAR-100, full-model tuning is over 3\% higher at $|M|=2000$. We also stress that the residual adapters are still usable with high buffer capacity, improving nearly 8\% from $f_T$. For both datasets, we set the threshold $m=500$ for the practical algorithm to switch from residual adapters to full-model tuning.

\begin{table}[t]
    \centering
    \scriptsize
    \begin{tabular}{c|ccc|ccc}
        & \multicolumn{3}{c|}{\textbf{CIFAR-10}} & \multicolumn{3}{c}{\textbf{CIFAR-100}}\\
        \midrule
        $f_T$ Accuracy & \multicolumn{3}{c|}{$81.7 \pm 0$} & \multicolumn{3}{c}{$57.1 \pm 0$}\\
        \midrule
        \textbf{Adaptation Routine} & $|M|=200$ & $|M|=500$ & $|M|=2000$ & $|M|=200$ & $|M|=500$ & $|M|=2000$\\
        \midrule
        Residual Adapter & $\mathbf{84 \pm 0.2}$ & $\mathbf{86.3 \pm 0.2}$ & $89.0 \pm 0.1$ & $\mathbf{59.9 \pm 0.1}$ & $\mathbf{61.5 \pm 0.1}$ & $64.7 \pm 0.2$\\
        Full-model tuning & $81.5 \pm 0.03$ & $84.1 \pm 0.4$ & $\mathbf{89.4 \pm 0.2}$ & $56.5 \pm 0.3$ & $61.1 \pm 0.2$ & $\mathbf{68.0 \pm 0.2}$\\
    \end{tabular}
    \caption{Classification accuracy of different adaptation routines for \algname{}~(Ridge).}
    \label{tab:buffer_size}
\end{table}

\section{Discussion}
In this work, we introduced the notion of \textit{schedule-robustness} in CL, which requires algorithms to perform consistently across different data streams. We argued that such property is key to common CL applications, where the schedule cannot be chosen by the learner and is often unknown a priori. In contrast, most existing methods disregard this aspect and are often designed around specific schedules. We empirically demonstrated that this severely degrades their performance when a different schedule is used. To tackle the above issues, we proposed \algname{} and formulated CL as the problem of learning online a schedule-robust classifier, followed by predictor adaptation using only replay data. \algname{} is schedule-robust by design and empirically outperforms existing methods across all evaluated schedules, even those for which they were originally optimized. 

We now focus on two key aspects that emerged from our analysis, which we believe might provide useful insights into future work on CL:

\paragraph{Pre-training and CL} The results from our experiments show that pre-training is broadly beneficial for all evaluated methods (see our \cref{tab:main_comp} vs Tab. 2 in \cite{buzzega2020dark}). We believe that this observation should call for \textit{greater focus on how to exploit pre-training to develop robust CL methods}: our choice to pre-train the feature extractor $\embd$ in \algname{} and combining it with linear online methods such as incremental Ridge Regression, is a principled decision derived from our meta-learning-based perspective, which also helps to mitigate forgetting. In contrast, most previous CL methods adopting pre-training see it as a preliminary phase rather than an integral component of their methods~\citep{wang2022learning, wu2022class}. This approach often leads to sub-optimal design decisions and potentially less effective techniques for leveraging pre-training, as empirically observed in \cite{mehta2021empirical} and further highlighted by our experiments (see also \cref{sec:app_comp_ens} for the comparison between \algname{} (Ridge) and the ensemble model by \cite{shanahan2021encoders}).

Pre-training is also a realistic assumption. We have argued that application domains (e.g., computer vision and natural language understanding) have vast amounts of curated data and increasingly available pre-trained models that CL algorithms could readily exploit. Additionally, we demonstrated in \cref{sec:exp_mimg} that pre-training with even a modest dataset is very beneficial, which is a promising result for applying this strategy to other domains. In particular, we note these settings are analogous to initializing a CL algorithm with a random model but then having a large first batch (i.e., containing samples from several classes), a schedule that is gaining increasing traction in the literature \citep{hou2019learning, mittal2021essentials}. Our strategy extends naturally to these settings if we allow offline training on the first batch to replace pre-training.

\paragraph{Class-independent Online Learning and On-demand Adaptation}
We believe that \algname{} provides a new general practice to approach CL problems. In particular, the strategy of {\itshape first learning online a coarse-yet-robust model for each individual class and then adapting it on-demand on data of interest} grants flexibility as well as consistency to CL algorithms. The online step focuses on learning each class independently, which prevents inter-class interference~\citep{hou2019learning} by construction. This single-class learning process is open-ended (i.e. one-vs-all for each class) with any new classes easily assimilated. We have demonstrated that the predictor $f_T$ resulting from this first step, is already competitive with most previous CL algorithms. Then, at any moment $T$ in time (not necessarily at the end of the sequence), we can adapt $f_T$ to a $f^*$ specializing to all the observed classes so far. This step is performed with a cross-entropy loss, which enforces interaction among classes and ultimately leads to improved performance. \algname{}'s robust performance offer strong support to adopting this strategy more broadly in CL.  



\paragraph{Limitation and Future Works}
We close by discussing some limitations of our work and future research directions. Firstly, while model adaptation works well in practice, most previously observed data are omitted from directly updating the feature extractor. One key direction we plan to investigate is how to perform the periodic update of the feature extractor {\itshape during CL}, to improve data efficiency while preserving schedule-robustness. Secondly, we note that the residual adapters in our model are specific to convolutional networks. We will extend our method to more powerful architectures like transformers.



{
\bibliographystyle{unsrtnat}
\bibliography{biblio}
}

\newpage

\appendix

\crefname{assumption}{Assumption}{Assumptions}
\crefname{equation}{}{}
\Crefname{equation}{Eq.}{Eqs.}
\crefname{figure}{Figure}{Figures}
\crefname{table}{Table}{Tables}
\crefname{section}{Section}{Sections}
\crefname{theorem}{Theorem}{Theorems}
\crefname{proposition}{Proposition}{Propositions}
\crefname{fact}{Fact}{Facts}
\crefname{lemma}{Lemma}{Lemmas}
\crefname{corollary}{Corollary}{Corollaries}
\crefname{example}{Example}{Examples}
\crefname{remark}{Remark}{Remarks}
\crefname{algorithm}{Algorithm}{Algorithms}
\crefname{enumi}{}{}

\crefname{appendix}{Appendix}{Appendices}

\numberwithin{equation}{section}
\numberwithin{lemma}{section}
\numberwithin{proposition}{section}
\numberwithin{theorem}{section}
\numberwithin{corollary}{section}
\numberwithin{definition}{section}
\numberwithin{algorithm}{section}
\numberwithin{remark}{section}

\section*{\Huge\textbf{Appendix}}

This appendix is organized as follows:
\begin{itemize}
    \item In \cref{sec:app_ls} we comment on the schedule-robustness property of Ridge Regression;
    \item In \cref{sec:app-additional-exp} we present additional empirical investigation on the behavior of \algname{} and its main components; 
    \item In \cref{sec:app_model} we report the implementation details of \algname{} used in our experiments;  
    \item Finally, in \cref{sec:app_base_tune} we provide more information on model tuning and baseline methods.
\end{itemize}

\section{Ridge Regression and Schedule-robustness}
\label{sec:app_ls}
We show that Ridge Regression is schedule-robust and mitigates forgetting. From \cref{sec:method_ci_meta}, the Ridge Regression classifier is defined as $f(x) = \argmax_{y}~ w_y^\top\embd(x)$, where
\eqal{\label{eq:app_ridge_reg}
     W^* = [w_1 \dots w_K] = \argmin_{W}\frac{1}{|D|}\sum_{(x,y)\in D}~~\nor{W^\top\embd(x) - \textrm{OneHot}(y)}^2 + \lambda\nor{W}^2,
}
with $\embd$ as a feature extractor. 
The closed-form solution for $W$ is
\eqal{
    W^* & = (X^\top X + \lambda I)^{-1}X^\top Y \qquad \textrm{where}\label{eq:app_ridge_sol}\\
     & X = [\embd(x_1) \dots \embd(x_N)]^\top \textrm{ and } Y=[\textrm{OneHot}(y_1) \dots \textrm{OneHot}(y_N)]^\top.\label{eq:app_ridge_var}
}
Substituting \cref{eq:app_ridge_var} into \cref{eq:app_ridge_sol}, we have
\eqal{
\label{eq:app_seq_ridge}
    X^\top X = \sum_{i=1}^N\embd(x_i)\embd(x_i)^\top \textrm{ and } X^\top Y = [c_1 \dots c_K],
}
where $c_i = \sum_{x\in X^i}\embd(x)$ is the sum of embeddings from class $X^i$, and $K$ is the total number of classes in the dataset $\D$.

From \cref{eq:app_seq_ridge}, it is clear that both $X^\top X$ and $X^\top Y$ are invariant to the ordering of samples in $\D$. Therefore computing $W^*$ is also invariant to sample ordering in $\D$. We thus conclude that Ridge Regression is schedule-robust. Since Ridge Regression learns each $w_i$ independently, it also mitigates catastrophic forgetting caused by class interference.

\section{Additional Experiments}
\label{sec:app-additional-exp}
\subsection{Additional Results on Different Splits for CIFAR Datasets}
\label{sec:app_add_split}
We compare \algname{} with the baseline methods on CIFAR-10 10-split and CIFAR-100 50-split respectively. The classification accuracies are reported in \cref{tab:app_main_comp_cifar2class}.
\begin{table}[htb]
    \centering
    \scriptsize
    \begin{tabular}{ll|ccc|ccc}
        & & \multicolumn{3}{c|}{\textbf{CIFAR-10 10-split}} & \multicolumn{3}{c}{\textbf{CIFAR-100 50-split}}\\
        \midrule
        \multicolumn{2}{c|}{Joint Training (i.i.d.)} & \multicolumn{3}{c|}{$94.2$ \tiny{$\pm 1.6$}} & \multicolumn{3}{c}{$79.3$ \tiny{$\pm 0.4$}}\\
        \midrule
        \multicolumn{2}{c|}{\textbf{CL Algorithm}} & $|M|=200$ & $|M|=500$ & $|M|=2000$ & $|M|=200$ & $|M|=500$ & $|M|=2000$\\
        \midrule
        \multirow{4}{*}{\rotatebox[origin=c]{90}{\textsc{Offline}}} 
        & ER    & $54.8$ \tiny{$\pm 1.8$} & $71.5$ \tiny{$\pm 1.1$} & $85.8$ \tiny{$\pm0.1$} & $15.1$  \tiny{$\pm 2.1$} & $ 28.5$ \tiny{$\pm 1.1$} & $ 50.3$ \tiny{$\pm 0.4$} \\
        & BIC & $64.9$ \tiny{$\pm 2.0$} & $78.8$ \tiny{$\pm 0.5$}  & $87.5$ \tiny{$\pm0.6$}  & $16.1$ \tiny{$\pm 4.1$} & $25.7$ \tiny{$\pm 0.6$} & $36.8$ \tiny{$\pm 2.7$} \\
        & DER   & $54.4$ \tiny{$\pm 5.2$} & $63.2$ \tiny{$\pm 2.1$} & $75.4$ \tiny{$\pm7.4$} & $15.7$ \tiny{$\pm 1.7$} & $ 29.6$ \tiny{$\pm 2.0$} & $ 48.2$ \tiny{$\pm 1.4$} \\
        & DER++ & $62.9$ \tiny{$\pm 3.1$} & $68.8$ \tiny{$\pm 2.6$} & $77.4$ \tiny{$\pm5.8$} & $18.2$ \tiny{$\pm 1.4$} & $ 33.2$ \tiny{$\pm 2.0$} & $ 53.1$ \tiny{$\pm 0.9$} \\
        \midrule
        \multirow{5}{*}{\rotatebox[origin=c]{90}{\textsc{Online}}}
        & GDumb  & $73.0$ \tiny{$\pm 1.7$} & $81.1$ \tiny{$\pm 0.3$} & $87.1$ \tiny{$\pm 0.1$} & $22.8$    \tiny{$\pm 1.9$} & $37.1$ \tiny{$\pm 1.1$} & $58.9$ \tiny{$\pm 0.4$} \\
        & ER-ACE & $69.8$ \tiny{$\pm 1.8$} & $77.3$ \tiny{$\pm 0.8$} & $81.7$ \tiny{$\pm 1.2$} & $26.8$ \tiny{$\pm 0.9$} & $35.1$ \tiny{$\pm 0.9$} & $47.2$ \tiny{$\pm 0.5$}\\
        & ER-AML & $72.8$ \tiny{$\pm 1.7$} & $80.0$ \tiny{$\pm 0.3$} & $85.2$ \tiny{$\pm 0.6$} & $13.2$ \tiny{$\pm 1.1$} & $24.4$ \tiny{$\pm 1.8$} & $41.1$ \tiny{$\pm 1.1$}\\
        & SSIL & $65.1$ \tiny{$\pm 1.6$} & $70.7$ \tiny{$\pm 0.4$} & $74.2$ \tiny{$\pm 1.0$} & $32.9$ \tiny{$\pm 0.6$} & $40.1$ \tiny{$\pm 0.9$} & $52.2$ \tiny{$\pm 0.9$}\\
        & MIR & $52.6$ \tiny{$\pm 2.5$} & $63.8$ \tiny{$\pm 1.5$} & $72.5$ \tiny{$\pm 1.7$} & $7.9$  \tiny{$\pm 0.7$} & $17.1$ \tiny{$\pm 0.7$} & $41.1$ \tiny{$\pm 0.5$}\\
        \hdashline[1pt/1pt]\noalign{\vskip 1.0ex}
        \multirow{2}{*}{\rotatebox[origin=c]{90}{(ours)}}
        & \algname{} (NCC)   & $81.5$ \tiny{$\pm 0.1$} & $84.6$ \tiny{$\pm 0.1$} & $88.4$ \tiny{$\pm 0.1$} & $48.3$ \tiny{$\pm 0.1$} & $56.4$ \tiny{$\pm 0.2$} & $66.6$ \tiny{$\pm 0.2$} \\
        & \algname{} (Ridge) & $\mathbf{84}$ \tiny{$\pm 0.2$} & $\mathbf{86.3}$ \tiny{$\pm 0.2$} & $\mathbf{89.4}$ \tiny{$\pm 0.2$} & $\mathbf{59.9}$ \tiny{$\pm 0.1$} & $\mathbf{61.5}$ \tiny{$\pm 0.1$} & $\mathbf{68.0}$ \tiny{$\pm 0.2$}\\
    \end{tabular}
        
    \caption{Class-incremental classification accuracy on CIFAR-100 50-split and CIFAR-10 10-split.}
    \label{tab:app_main_comp_cifar2class}
\end{table}

\algname{}~(Ridge) outperforms all baselines in \cref{tab:app_main_comp_cifar2class}. We observe that most baseline methods' performance drop drastically compared to \cref{tab:main_comp} except GDumb, suggesting worse forgetting with more dynamic schedules, even though the underlying data remains unchanged. In contrast, \algname{} is able to perform consistently due to its schedule-robust design. The drop in baselines performance is also consistent with \cite{boschini2022transfer}: most existing methods are not designed to effectively exploit pre-training. When the same dataset is split into more tasks (or batches), the later ones benefit less from the prior knowledge encoded in the initial predictor $f_0$.

\subsection{Comparison with Ensemble Method}
\label{sec:app_comp_ens}
\begin{figure}[t]
    \centering
    \includegraphics[width=0.8\textwidth]{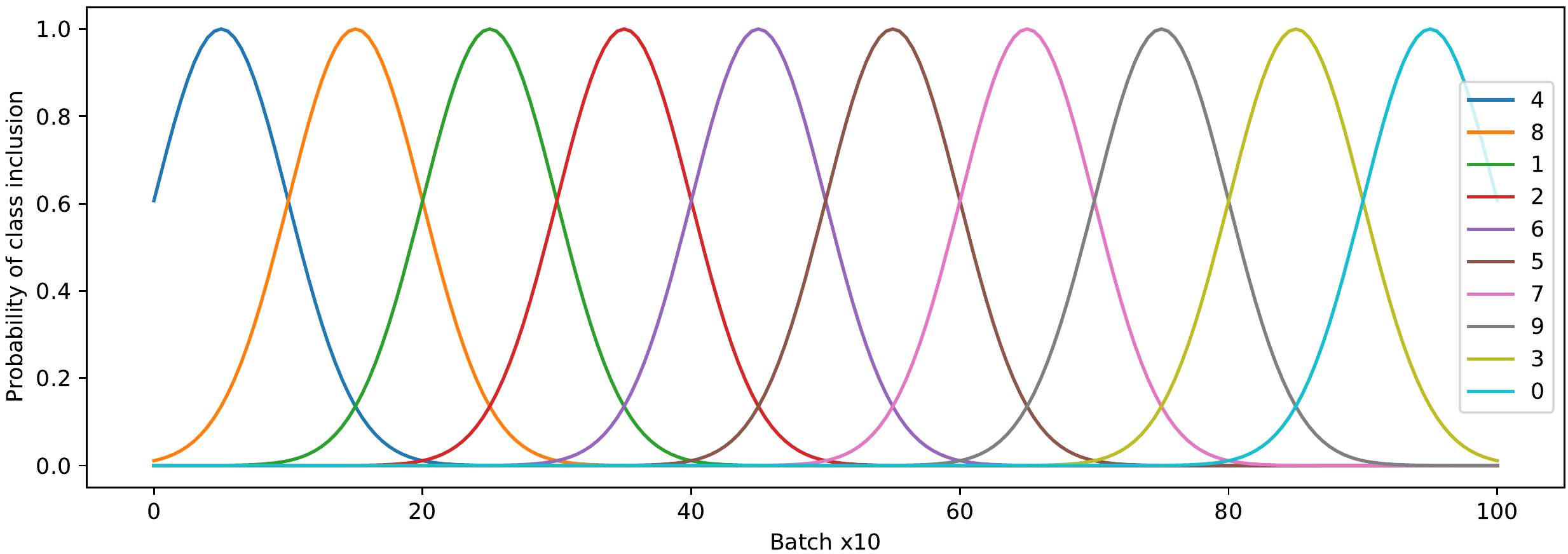}
    \caption{Example of Gaussian schedule. Figure originally from \cite{shanahan2021encoders}.
    The schedule proposes to sample a class with a Gaussian probability, where each class peaks at a different time. There are no class boundaries, and the probability of sampling each class evolves smoothly over time.
    }
    \label{fig:gaussian_schedule_ENS}
\end{figure}

We also compare \algname{} to ENS~\citep{shanahan2021encoders}, which is designed to tackle different schedules, including a Gaussian schedule introduced in the same paper (see ~\cref{fig:gaussian_schedule_ENS}). Specifically, ENS keeps the pre-trained feature extractor fixed and learns an ensemble of classifiers for the novel classes. For ENS, we keep its original design for pre-training: using ResNet50 via self-supervised learning on ImageNet~\footnote{ResNet50 checkpoint available at \url{https://github.com/deepmind/deepmind-research/tree/master/byol}}. We also include the memory-free variant of \algname{}~(Ridge) in the comparison, since ENS does not use ER.

\begin{table}[htb]
    \centering
    \scriptsize
    \begin{tabular}{c|ccc|ccc}
        & \multicolumn{3}{c|}{\textbf{CIFAR-10}} & \multicolumn{3}{c}{\textbf{CIFAR-100}}\\
        \toprule
        \textbf{CL Algorithms} & \textbf{5-split} & \textbf{10-split} & \textbf{Gaussian schd} & \textbf{20-split} & \textbf{100-split} & \textbf{Gaussian Schd}\\
        \midrule
        ENS (ResNet50) & $79.0$ \tiny{$\pm 0.4$} & $78.3$ \tiny{$\pm 0.4$} & $50.1$ \tiny{$\pm 9.5$} & $55.3$ \tiny{$\pm 0.4$} & $54.1$ \tiny{$\pm0.5$} & $39.0$ \tiny{$\pm 1.4$}\\
        \algname{} (ResNet50, $|M|=0$) & \multicolumn{3}{c|}{$\mathbf{88.1}$ \tiny{$\pm 0$}} & \multicolumn{3}{c}{$\mathbf{67.2}$ \tiny{$\pm 0$}}\\
        \hdashline[1pt/1pt]\noalign{\vskip 1.0ex}
         \algname{} (ResNet18, $|M|=0$) & \multicolumn{3}{c|}{$81.7$ \tiny{$\pm 0$}} & \multicolumn{3}{c}{$57.1$ \tiny{$\pm 0$}}\\
         \algname{} (Resnet18, $|M|=500$) & \multicolumn{3}{c|}{$86.3$ \tiny{$\pm 0.2$}} & \multicolumn{3}{c}{$61.5$ \tiny{$\pm 0.1$}}\\
    \end{tabular}
    \caption{Classification accuracy comparison \algname{}~(Ridge) vs. Ensemble.}
    \label{tab:app_main_comp_ens}
\end{table}

In \cref{tab:app_main_comp_ens}, we observe that the ResNet50 provided by ENS is substantially more powerful than the ResNet18 used in our main experiments. In addition, memory-free \algname{} returns deterministic predictors $f_T$ for all possible schedules, while the performance of ENS degrades on the Gaussian schedule, despite its robustness in handling different splits. The results further validates that it is intractable to optimize CL algorithms against all schedules, and schedule-robustness should be considered as part of algorithm design. Lastly, we note that \algname{} has the additional advantage of using ER to further improve performance, compared to the lack of ER in ENS.

\subsection{The Effects of Buffering Strategy}
\label{sec:app_buffer}
In this section, we study different buffering strategies used in ER. We consider both exemplar selection~\citep{rebuffi2017icarl} and random reservoir sampling~\citep{vitter1985random}, which are commonly used in existing works. We also introduce two simple variants to exemplar selection, including selecting samples nearest to the mean representation of each class (Nearest Selection), or selecting samples furthest to the mean representation of each class (Outlier Selection).

\begin{table}[htb]
    \footnotesize
    \centering
    \begin{tabular}{c|cc|cc}
         & \multicolumn{2}{c|}{\textbf{CIFAR-10}} & \multicolumn{2}{c}{\textbf{CIFAR-100}} \\
         \midrule
         \textbf{Buffering strategy} & $|M|=500$ & $|M|=2000$ & $|M|=500$ & $|M|=2000$\\
         \midrule
         Exemplar & $86.3$ \tiny{$\pm 0.2$} & $89.4$ \tiny{$\pm 0.2$} & $61.5$ \tiny{$\pm 0.1$} & $68.0$ \tiny{$\pm 0.2$}\\
         Random & $84.5$ \tiny{$\pm 0.2$} & $88.8$ \tiny{$\pm 0.1$} & $61.1$ \tiny{$\pm 0.1$} & $65.4$ \tiny{$\pm 0.1$}\\
         Nearest & $72.3$ \tiny{$\pm 0.1$} & $82.2$ \tiny{$\pm 0.1$} & $55.6$ \tiny{$\pm 0.1$} & $57.0$ \tiny{$\pm 0.2$}\\
         Outlier & $72.6$ \tiny{$\pm 0.4$} & $82.1$ \tiny{$\pm 0.3$} & $55.5$ \tiny{$\pm 0.1$} & $57.0$ \tiny{$\pm 0.3$}\\
    \end{tabular}
    \caption{Classification accuracy with different buffering strategy}
    \label{tab:exemplar}
\end{table}

We report the effects of different buffering strategies in \cref{tab:exemplar}. The exemplar strategy clearly performs the best, followed by random reservoir sampling. Both nearest and outlier selection perform drastically worse. From the results, we hypothesize that an ideal buffering strategy should store representative samples of each class (e.g., still tracking the mean representation of each class as in exemplar selection), while maintaining sufficient sample diversity (e.g., avoid nearest selection). We remark that the results in \cref{tab:exemplar} are obtained using a fixed feature extractor $\embd$. We leave it to future work to investigate the setting where $\embd$ is continuously updated during CL.

The results have several implications for CL algorithms. While random reservoir sampling performs well \textit{on average}, it can perform poorly in the \textit{worst cases}, such as when the random sampling happens to coincide with either nearest or outlier selection. Therefore, it appears helpful to rank the samples observed during CL and maintain representative yet diverse samples. On the other hand, the results show that GDumb~\citep{prabhu2020gdumb} is not fully schedule-robust, since random reservoir sampling would perform poorly in worst cases.

\subsection{Buffering Strategy and Schedule-Robustness}
\label{sec:app_schd}

As discussed in \cref{sec:method_replay}, exemplar selection yields a deterministic replay buffer when one or more classes are presented in a single batch. In this section, we further investigate its behaviors under schedules with smaller batches, where data from a class is spread out in a sequence. We use random reservoir sampling as a baseline for comparison.

In this experiment, we consider each class separately, since both exemplar strategy and random sampling select samples from each class independently. We thus use buffer size $b_1$ and batch size $b_2$ on a per-class basis here. Batch size refers to the number of samples in a batch for a single class, while buffer size refers to the allowed buffer capacity per class.

We use the $\ell_2$ distance $\nor{\hat\mu_y - \mu_y^*}$ as the metric to measure how well the stored samples in the last replay buffer $M_T$ approximates the population mean of each class, where $\hat\mu_y$ is the mean of the points in $M_T$ belonging to class $y\in\Y$ and $\mu^*_y$ the mean of $X^y$, namely all points belonging to class $y$ in the dataset $\D$.

We consider 4 scenarios: 1) $b_1=20, b_2=20$; 2) $b_1=20, b_2=80$; 3) $b_1=90, b_2=10$; 4) $b_1=50, b_2=50$. We also note that $b_1=20$ represents one of the worst possible scenario for CIFAR-10, since it implies a total buffer size of 200 and no class is allowed to temporarily borrow capacity from other classes. All scenarios are run with 100 random shuffling of the class data to simulate various schedules.

\begin{figure}[t]
    \centering
    \includegraphics[width=\textwidth]{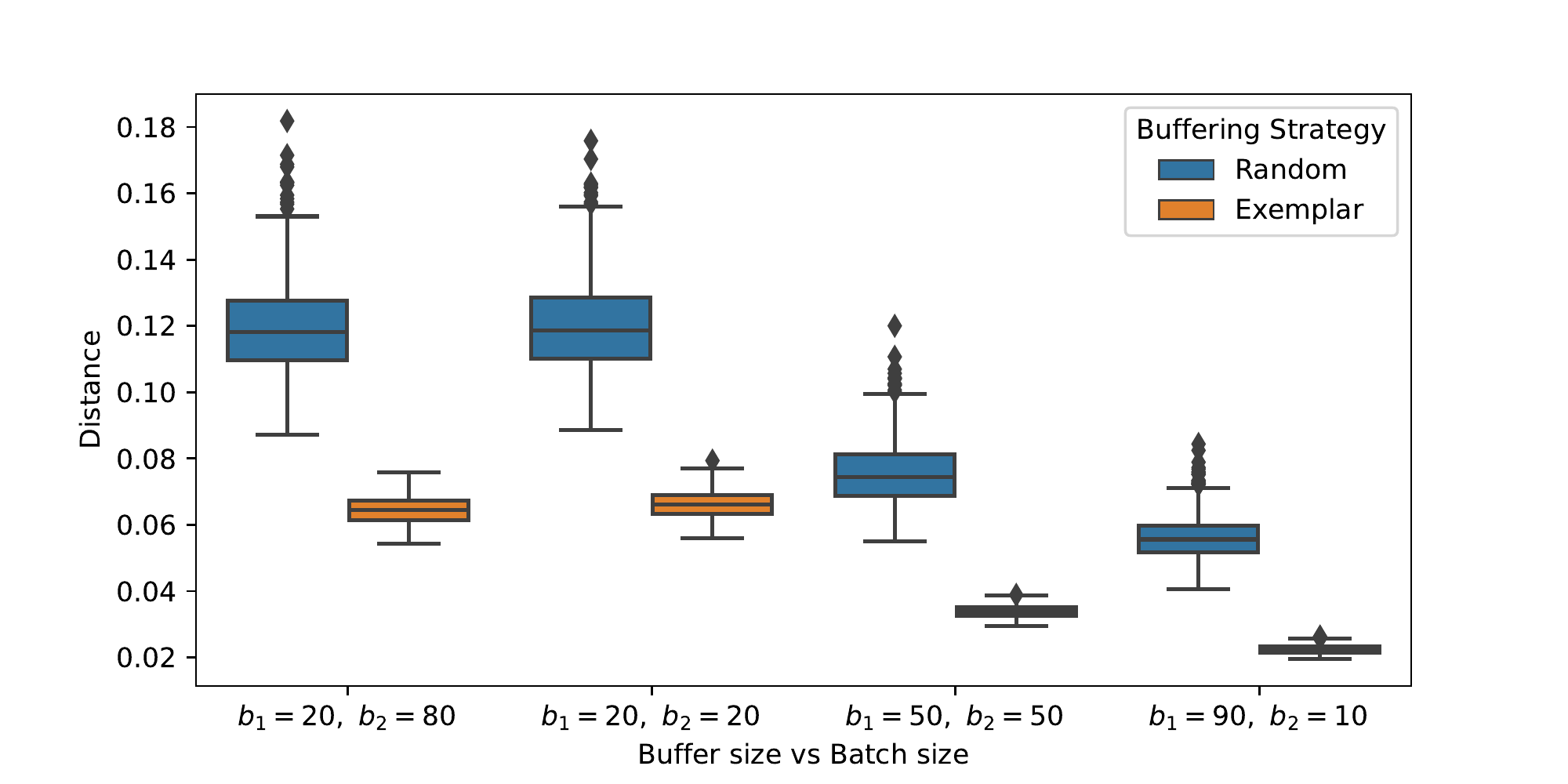}
    \caption{Effects of buffer size and batch size on selected samples in $M_T$.  Exemplar selection vs. Random Sampling. Lower distance and variance are better. Distance denotes how well the selected samples approximates the population mean. Lower variance suggests similar distribution at $M_T$.}
    \label{fig:icarl_vs_random}
\end{figure}

\cref{fig:icarl_vs_random} shows that the exemplar strategy is much better at approximating the population mean of each class for all scenarios, compared to random sampling. This corroborates our results from \cref{sec:app_buffer} that exemplar selection selects more representative samples and thus performs better during ER. In addition, we note that exemplar selection achieves drastically lower variance for all settings, suggesting that the selected samples at $M_T$ form similar distributions and make \algname{} schedule-robust.

The experiment also suggests practical ways to ensure schedule-robustness. We first observe that the exemplar strategy decide on what samples to keep based on the current batch and the current buffer $M_t$. Therefore its behavior is predominantly determined by the combined size of $b_1$ and $b_2$. Consequently, simply use a large buffer (e.g. $b_1=90$ in scenario 4) will yield a $M_T$ closely approximating the desired population mean, even when the batch size $b_2=10$ is very small and generally considered as challenging for CL.

\section{Model Details}
\label{sec:app_model}
For CIFAR datasets, we use a ResNet18 backbone pre-trained on the Meta-Dataset~\citep{triantafillou2019meta}. The model definition and pre-trained representation can be found here\footnote{ \url{https://github.com/VICO-UoE/URL}}. 
For \mimg{}, we use a ResNet-12 backbone as commonly adopted in existing meta-learning works~\citep{lee2019meta, oreshkin2018tadam, ravichandran2019few, tian2020rethinking}, and use the default architecture from the official implementation of \cite{tian2020rethinking}. Please refer to the accompanying source code for more details.


Following \cite{tian2020rethinking}, we find it beneficial to normalize the embedding $\embd(x)$ to unit length before classification. The normalization is implemented in predictor $f_t$ as a non-linearity layer.

\paragraph{Adapting NCC with Experience Replay} Since we normalize sample embedding, NCC can be implemented as a standard linear layer. Recall that NCC is
\eqal{\label{eq:app_proto}
    f(x) = \argmin_{y} \nor{\embd(x) - c_y}^2_2 \quad \textrm{ s.t. } \quad c_y = \frac{1}{n}\sum_{x\in X^y} \embd(x).
}
When $\embd(x)$ is unit-length,
\eqal{\label{eq:app_proto_linear}
    \argmin_{y} \nor{\embd(x) - c_y}^2_2 & = \argmin_{y} ~~\embd(x)^\top \embd(x) -\embd(x)^\top c_y + c_y^\top c_y\\
    & = \argmin_{y} ~~ 1 - \embd(x)^\top c_y + c_y^\top c_y\\
    & = \argmax_{y} ~~ \embd(x)^\top c_y - c_y^\top c_y.
}
We thus use $c_y$ and $- c_y^\top c_y$ to initialize the weights and bias for class $y$ respectively. Adapting NCC with ER therefore treats NCC as a standard linear layer as described above.

\paragraph{Optimizer} Following \cite{li2022cross}, we use AdaDelta~\citep{zeiler2012adadelta} for model adaptation with ER.

\paragraph{ER Hyper-parameters} We list the hyper-parameters for ER in \cref{tab:hyper-params} and state the values used in our grid search.

\begin{table}[htb]
    \centering
    \begin{tabular}{c|cc}
        \textbf{Hyper-parameter} & \textbf{Description} & \textbf{Possible values} \\
        \midrule
        $l_1$ & learning rate for classifier $\ridge$ & [0.5, 0.1, 0.01, 0.001]\\
        $l_2$ & learning rate for feature extractor $\embd$ & [0.02, 0.01, 0.002, 0.001]\\
        $\tau$ & temperature for cross-entropy loss & [1, 2, 4, 5]\\
        $b$ & batch size from ER & [50, 100]\\
    \end{tabular}
    \caption{ER Hyper-parameters in the proposed method and their values used in the grid search.}
    \label{tab:hyper-params}
\end{table}

\section{Model Tuning for Baseline Methods}
\label{sec:app_base_tune}
\subsection{CIFAR-10 / 100}

\paragraph{Dataset} We resize the input images to $84\times84$, both at training and test time, to be compatible with the pre-trained representation from~\cite{li2021universal}. During training, we apply random cropping and horizontal flipping as standard data augmentations.

\paragraph{Hyperparameters}
Starting from the exhaustive search in \citet{buzzega2020dark}, we tuned all the baselines by testing different learning rates and method-specific hyperparameters using the official repository\footnote{ \url{https://github.com/aimagelab/mammoth}}. We used a vanilla SGD optimizer without momentum and batch size of $32$, training for $50$ epochs per task for all offline algorithms. We found that a good trade-off between preserving and updating the knowledge in the pre-trained representation is given by learning rates of $0.01$ and $0.03$. For small buffer sizes, typically $lr=0.01$ achieves a lower standard deviation across different random seeds at the cost of slightly lower average performance, while $lr=0.03$ works well with larger memory. Other empirical works also recommend updating the model parameters with lower learning rates by studying the training dynamics of CL algorithms~\citep{mirzadeh2020understanding}.
For \textbf{ER} and \textbf{BIC}~\cite{wu2019large}, we used $lr=0.01$ (no other hyperparameters are present).
For \textbf{DER/DER++}~\citep{buzzega2020dark}, we tested different values of replay loss coefficients $\alpha, \beta \in \{0.1, 0.3, 0.5, 1.0 \}$, defined in Eq.6 of the original paper. While for small buffer sizes ($200$, $500$), no clear winner emerges, we found that larger buffer sizes ($2000$) benefit from larger coefficients. Assuming that the actual data schedule is not known in advance, safe choices for $\alpha$ are $0.5$ or $0.3$, confirming the original paper suggestion. DER++ usually requires $\beta \ge 0.5$ and is less sensitive to the choice of $\alpha$. For CIFAR-100 $10$-split we used $lr=0.01$, $\alpha=0.5$, $\beta=0.5$, while for the $50$-split $lr=0.03$ works best.

All online baselines use a single epoch for training. For \textbf{GDumb}~\citep{prabhu2020gdumb}, we decay $lr$ to $5\times10^{-4}$ with Cosine annealing and modify the original training recipe by removing CutMix regularization~\citep{yun2019cutmix} as we found it not to be compatible with the pre-trained representation. For other online methods, we follow the official implementation\footnote{\url{https://github.com/pclucas14/aml}} of \textbf{ER-AML} and \textbf{ER-ACE}~\citep{caccia2022new} and tunes the learning rate of each method. We observe similarly that $lr=0.01$ works well for all. Loss coefficients are set using the recommended values in the original implementation.

Moreover, our experiments confirm that, if an algorithm is not schedule-robust, its hyper-parameters should be tuned independently for each schedule. However, this is infeasible as the actual schedule is generally unknown, further motivating the necessity of designing schedule-robust CL algorithms.

%

\subsection{ \mimg{}}

\paragraph{Dataset} For \mimg{}, we used the $64$ classes of the meta-training split for pre-training the representation. Then, we train and test all CL algorithms on the $20$ meta-test classes, by splitting the dataset in $10$ sequential binary tasks. During training, we apply random cropping and horizontal flipping as standard data augmentations. Input images have shape $84\times84\times3$.

\paragraph{Hyperparameters}
Similarly as for CIFAR-10/100, we search for the best hyperparameters of the baseline methods. We train all algorithms with SGD optimizer without momentum and batch size of $32$. We found that for buffer size $|M| = 200$, training for $10$ epochs is enough, while for $|M|=500$, better performance are obtained with $20$ epochs. Best hyperparameters are reported below.

\begin{itemize}[\null,leftmargin=*]
    \item \textbf{Offline Methods}
    \item \textit{Buffer size} $|M| = 200$~~($10$ epochs)
    \begin{itemize}
        \item \textbf{ER}: $lr=0.01$
        \item \textbf{BIC}: $lr=0.01$
        \item \textbf{DER}: $lr=0.03,~\alpha=1.0$
        \item \textbf{DER++}: $lr=0.03,~\alpha=1,~\beta=1.0$
    \end{itemize}
    \item \textit{Buffer size} $|M| = 500$~~($20$ epochs)
    \begin{itemize}
        \item \textbf{ER}: $lr=0.01$
        \item \textbf{BIC}: $lr=0.01$
        \item \textbf{DER}: $lr=0.05,~\alpha=1.0$
        \item \textbf{DER++}: $lr=0.05,~\alpha=0.5,~\beta=1.0$
    \end{itemize}

    \item \textbf{Online Methods}
    \item \textit{Buffer size} $|M| = 200$ ~~($1$ epoch)
    \begin{itemize}
        \item \textbf{GDumb}: $lr=0.03$ decayed to $5\times10^{-4}$ with Cosine annealing, ER epoch $E=10$, No Cutmix.
        \item \textbf{ER-ACE}: $lr=0.01$
        \item \textbf{ER-AML}: $lr=0.01$, supercon temperature $\tau=0.2$
        \item \textbf{SSIL}: $lr=0.01$, distillation coefficient $\alpha=1$
        \item \textbf{MIR}: $lr=0.01$, sub-sampling rate $r=50$.
    \end{itemize}
    \item \textit{Buffer size}  $|M| = 500$ ~~($1$ epoch)
    \begin{itemize}
        \item \textbf{GDumb}: $lr=0.03$ decayed to $5\times10^{-4}$ with Cosine annealing, ER epoch $E=20$. No Cutmix.
        \item \textbf{ER-ACE}: $lr=0.01$
        \item \textbf{ER-AML}: $lr=0.01$, supercon temperature $\tau=0.2$
        \item \textbf{SSIL}: $lr=0.01$, distillation coefficient $\alpha=1$
        \item \textbf{MIR}: $lr=0.01$, sub-sampling rate $r=50$.
    \end{itemize}
\end{itemize}

\

\end{document}